\documentclass[runningheads]{llncs}
\usepackage{amsmath} 
\usepackage{booktabs} 
\usepackage{xcolor}
\usepackage{authblk} 
\usepackage{marvosym}
\usepackage{graphicx}
\usepackage[T1]{fontenc}
\usepackage{graphicx}
\begin{document}
\title{PX2Tooth: Reconstructing the 3D Point Cloud Teeth from a Single Panoramic X-ray}

\author{Wen Ma\footnote{Part of this work was done when Wen Ma was an intern at ChohoTech.}\inst{1,2}\and
Huikai Wu\inst{2}\and
Zikai Xiao\inst{1}\and
Yang Feng\inst{3}\and
Jian Wu\inst{1}\and
Zuozhu Liu\inst{1}\textsuperscript{(\Letter)} 
}
\authorrunning{Wen Ma et al.}
\institute{Zhejiang University, Zhejiang, China \and
Hangzhou ChohoTech, Zhejiang, China
\and
Angelalign Technology Inc., Shanghai, China\\
\email{zuozhuliu@zju.edu.cn}}
\maketitle 
\begin{abstract}
Reconstructing the 3D anatomical structures of the oral cavity, which originally reside in the cone-beam CT (CBCT), from a single 2D Panoramic X-ray(PX) remains a critical yet challenging task, as it can effectively reduce radiation risks and treatment costs during the diagnostic in digital dentistry. However, current methods are either error-prone or only trained/evaluated on small-scale datasets (less than 50 cases), resulting in compromised trustworthiness. In this paper, we propose PX2Tooth, a novel approach to reconstruct 3D teeth using a single PX image with a two-stage framework. First, we design the PXSegNet to segment the permanent teeth from the PX images, providing clear positional, morphological, and categorical information for each tooth. Subsequently, we design a novel tooth generation network (TGNet) that learns to transform random point clouds into 3D teeth. TGNet integrates the segmented patch information and introduces a Prior Fusion Module (PFM) to enhance the generation quality, especially in the root apex region. Moreover, we construct a dataset comprising 499 pairs of CBCT and Panoramic X-rays. Extensive experiments demonstrate that PX2Tooth can achieve an Intersection over Union (IoU) of 0.793, significantly surpassing previous methods, underscoring the great potential of artificial intelligence in digital dentistry.


\end{abstract}

\section{Introduction}
Cone-Beam Computed Tomography (CBCT)~\cite{ref_article17} provides detailed volumetric three-dimensional information of anatomical structures, playing a critical role in dental treatment such as implanting and orthodontics~\cite{ref_article24}. Despite its advantages, CBCT involves a certain amount of radiation doses and costs, which impedes its utility in the real world. In contrast, Panoramic X-ray (PX) images, characterized by the low radiation exposure and cost-effective capture~\cite{ref_article25}, allow patients to undergo subsequent treatment without CBCT in many clinical scenarios, reducing both radiation exposure and economic burden~\cite{ref_article23}. Furthermore, PX imaging significantly enhances the efficiency of dental practices, both in preoperative and postoperative phases ~\cite{ref_article18}, marking a substantial leap toward digitalization in dentistry ~\cite{ref_article19}. 
\begin{figure}[ht!]
    \vspace{-2mm}
    \centering
    \includegraphics[width=1\linewidth]{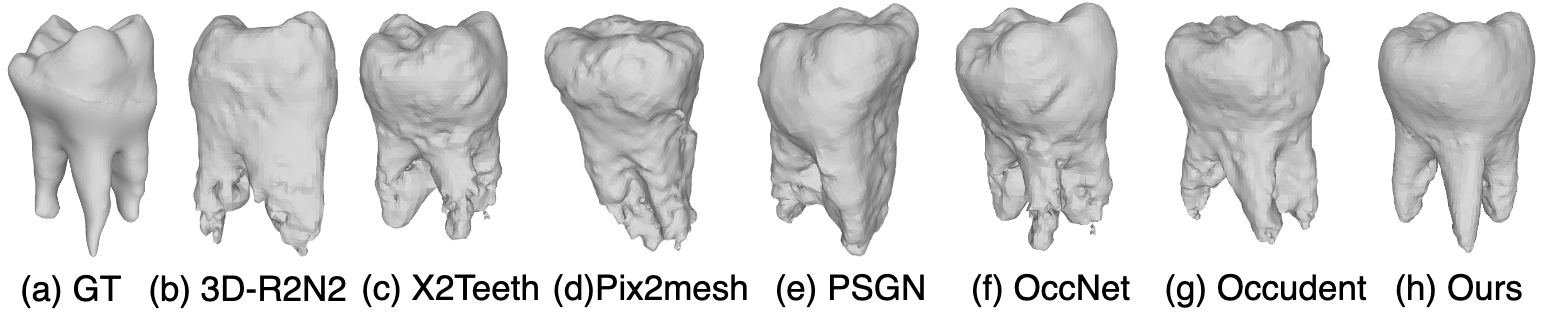}
    \caption{Generation quality illustration on a single tooth. PX2Tooth (ours) generates more accurate meshes representing tooth shapes, with significant improvements in the root tip area and the smoothness of the mesh surface.}
    \label{fig:visualize-single}
    \vspace{-5mm}
\end{figure}
This shift improves access to diagnostic and treatment options for a wider patient base. Nevertheless, while PX images provide 2D information about the oral cavity, they fall short in providing adequate 3D information regarding tooth volumes and spatial positioning, representing considerable difficulties yet holding substantial potential for advancing dental care~\cite{ref_article16}.

Recent research has launched attempts to reconstruct the 3D anatomical structures, such as teeth, in CBCT from the 2D PX images. These approaches fall into two categories. The first line of work reconstructed 3D teeth by augmenting PX images with additional labeling data, such as tooth landmarks or photographs of tooth crowns~\cite{ref_article25}. For instance,~\cite{ref_article7} developed a model that utilized tooth landmarks identified in PX images to predict tooth shapes. Similarly,~\cite{ref_article11} reconstructed an individual tooth by employing a shape prior and a reflectivity model derived from the corresponding crown photograph. These methods necessitate substantial time and effort from professional dentists for annotations, resulting in a slow, costly, and resource-intensive process. Another line of methods reconstructs 3D teeth only with a single PX image. For instance, existing works first segment the 2D teeth and subsequently employ generative networks such as GANs to reconstruct the 3D tooth structures ~\cite{ref_article10,ref_article14,ref_article6} Recent work also introduces a framework for  3D tooth reconstruction from PX images using a neural implicit function~\cite{ref_article12}. 

Nevertheless, there remain significant challenges for 3D tooth reconstruction from 2D PX images. First, the accuracy achieved by existing methods is still unsatisfactory to meet the requirements for clinical dental applications. Specifically, the detailed features in the root and the apex region of the tooth are inaccurate or inadequate. Moreover, these methods have only been validated on small datasets (comprising only 23-37 cases), while a large-scale dataset for more convincing evaluation is yet under development.

To address the aforementioned challenges, we introduce a novel method PX2Tooth which generates 3D point cloud teeth from single 2D PX images, eliminating the need for extra labeling while ensuring high precision. PX2Tooth works in a two-stage manner. First, we design a PXSegNet model to segment PX images into 32 permanent tooth categories, ensuring accurate morphological and categorical information for each tooth. The position and shape information of individual teeth are subsequently employed during the reconstruction process. 

Afterwards, we design a 3D tooth generation network TGNet, which transforms any 3D random point clouds to the desired teeth. The segmentation outputs from PXSegNet are treated as prior information, which is further integrated to guide the generation of TGNet with a Prior Fusion Module (PFM). Such prior information can help improve the generation quality, especially for the tooth root and tip areas. We also employ tailored segmentation, as well as reconstruction loss, functions to train PXSegNet and TGNet.

We evaluate the performance of our method with CBCT samples and corresponding generated PX images following existing works~\cite{ref_article10,ref_article6}. To ensure reliable assessment, we construct a dataset with 499 CBCT cases, which is one order of magnitude larger than existing datasets. Our experimental results show that PX2Tooth achieved a reconstruction Intersection over Union (IoU) of 0.793, which significantly surpasses existing state-of-the-art methods, as illustrated in Fig~\ref{fig:visualize-single}. Extensive analysis and ablation studies further reveal the effectiveness of our method, suggesting the great potential of AI in future digital dentistry.

\section{Methods}
\begin{figure}
    \vspace{-5mm}
    \centering
    \includegraphics[width=1\linewidth]{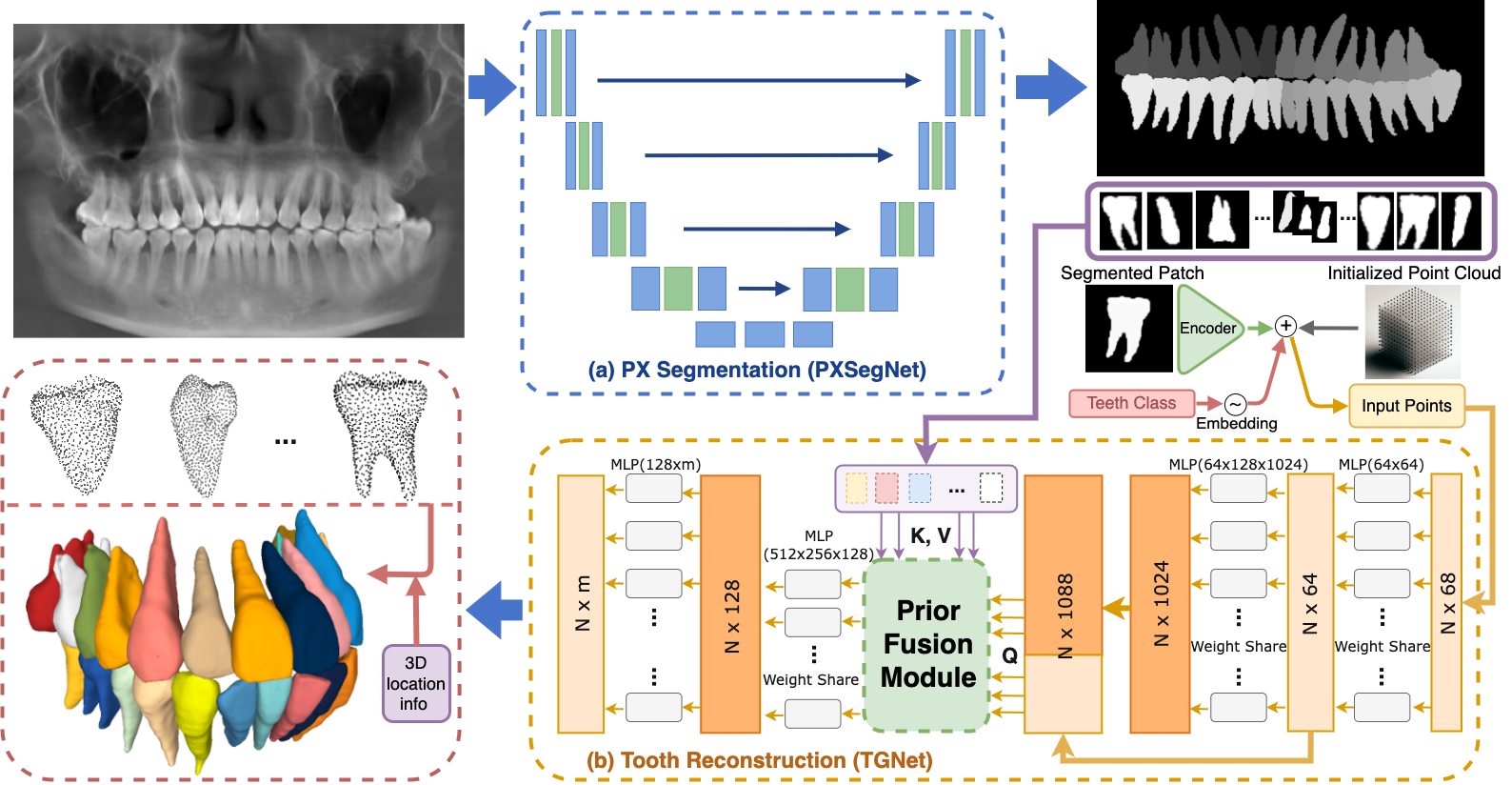}
    \caption{Pipeline of our PX2Tooth. (a) PXSegNet segments PX images into 32 teeth. The feature information and tooth categories of segmented patches are integrated into the initial point clouds. (b) TGNet generates individual teeth from the fused input points. A Prior Fusion Module(PFM) to provide more spatial morphological information for improved generation quality. Finally, the 3D registration matrix is used to restore the position of each tooth and obtain a real dental model.}
    \label{fig:PipeLine}
    \vspace{-5mm}
\end{figure}

PX2Tooth sequentially consists of two key components: the panoramic X-ray segmentation network (PXSegNet) and the Generative 3D tooth reconstruction network (TGNet). Based on the accurate and detailed information on the clear position and shape of each tooth obtained during the segmentation stage by PXSegNet, TGNet is capable of generating high-precision and smoothly contoured 3D point cloud representations of teeth. The overall architecture of PX2Tooth is shown in Fig.~\ref{fig:PipeLine}.

\subsection{2D Tooth Segmentation}

\subsubsection{PXSegNet.}

Using the entire PX image without segmentation to generate all 3D teeth would lead to clearer positional relationships among teeth, resulting in a more complicated reconstruction network, higher computational costs, or inferior accuracy.

Therefore, we propose PXSegNet, a Panoramic Segmentation Net that employs detailed segmentation technology to classify teeth within input PX images into 32 distinct categories aligned with the conventional FDI tooth numbering system, as illustrated in Fig.~\ref{fig}, thereby achieving precise dental modeling.

Given the potential for overlapping of adjacent teeth in the PX image, we accurately describe the 2D tooth segmentation as a multi-label segmentation problem~\cite{ref_article20}. The model takes an input PX image of size H×W. The segmentation output is of dimensions C×H×W, where the channel dimension C = 33 signifies the number of tooth classes: one for the background and 32 for individual teeth, akin to the approach in~\cite{ref_article9}. Consequently, the H×W output at each channel represents the segmentation output for each tooth class. These segmentation outputs are crucial in generating tooth patches for subsequent reconstruction, as elaborated upon in detail later. The PXSegNet architecture can be seen in Fig.~\ref{fig: PipeLine}(a), we leverage the effectiveness of UNet~\cite{ref_article12} as the foundational model. The notable advantage of UNet lies in its U-shaped structure, making it highly adept at accurate segmentation of medical images~\cite{ref_article22}.

To enhance the precision of segmenting small target teeth, our methodology integrates a dual-loss strategy,  encompassing Metric Boundary (MB) Loss and Unbalanced (UB) Loss. UB Loss excels in dealing with the global structure of the image and by introducing a factor $\gamma$ to prioritize challenging-to-segment samples, which helps solve the overall category imbalance. MB Loss calculates the intersection ratio of two sets to focus on the local similarity at the pixel level, which can improve the precision of segmenting the boundary of a single tooth.

Therefore, our method can effectively leverage their respective strengths to enhance overall accuracy.
The MB Loss and the UB Loss are defined as:

\begin{equation}
\text { MB Loss }=1-\frac{1}{C} \sum_{c=1}^{C} \frac{2 \sum_{i=1}^{N} y_{i, c} p_{i, c}}{\sum_{i=1}^{N} y_{i, c}+\sum_{i=1}^{N} p_{i, c}},
\end{equation}

\begin{equation}
\text { UB Loss }=-\frac{1}{N} \sum_{i=1}^{N} \sum_{j=1}^{C} y_{i, j}\left(1-p_{i, j}\right)^{\gamma} \log \left(p_{i, j}\right),
\end{equation}
where $C$ represents The number of teeth classes in the classification problem, and $N$ denotes the number of samples or instances in the dataset. $p$ is the predicted probability. $y$ represents The ground truth (GT)  label.

\subsection{3D Tooth Reconstruction}

\subsubsection{TGNet.} 

We introduced a novel point cloud Teeth Generative Net(TGNet) for 3D teeth reconstruction inspired by PointNet~\cite{ref_article13}. TGNet exhibits enhanced conciseness and flexibility in generating 3D point clouds from a PX image network. Based on PointNet, we enrich the input information, which includes segmented tooth patches, tooth categories, and initialized point clouds. Our approach aims to alter the output to generate the point cloud that accurately represents the position and fine shape of the teeth.

Furthermore, we tailored a reconstruction loss to better suit the new generation task of TGNet, as opposed to the default loss used in PointNet, which is more aligned with segmentation or classification algorithms.

The Reconstruction Loss (RT Loss) measures the similarity between the predicted and ground truth point clouds, 

which is defined as:
\begin{equation}
\text{RT Loss}(A, B) = \sum_{i=1}^{m} \min_{j=1}^{n} \|A_i - B_j\|^2 + \sum_{j=1}^{n} \min_{i=1}^{m} \|B_j - A_i\|^2,
\end{equation}
Where $B$ and $A$ denote the predicted point cloud of $m$ points and the ground truth (GT) point cloud of $n$ points, respectively.

\subsubsection{Prior Fusion Module.} To enhance the matching degree of the generated tooth apex and improve the uniform distribution of the generated point cloud, we propose a novel Prior Fusion Module (PFM). PFM integrates 3D features with 2D features to enhance the accuracy of the tooth tip(Fig.3B). The PFM utilizes the point cloud information (Q) as queries while employing the image information (K and V) as keys and values, respectively. Here, K and V represent the spatial information of individual teeth images and their high-dimensional feature information obtained in the segmentation process. 

The PFM mechanism enhances model focus on individual point cloud elements by linking them to corresponding image elements, enriching morphological and spatial representations. This module efficiently utilizes both point cloud and image data to refine the alignment of spatial and morphological features across modalities.

\begin{figure}[ht]
    \vspace{-5mm}
    \centering
    \includegraphics[width=0.8\linewidth]{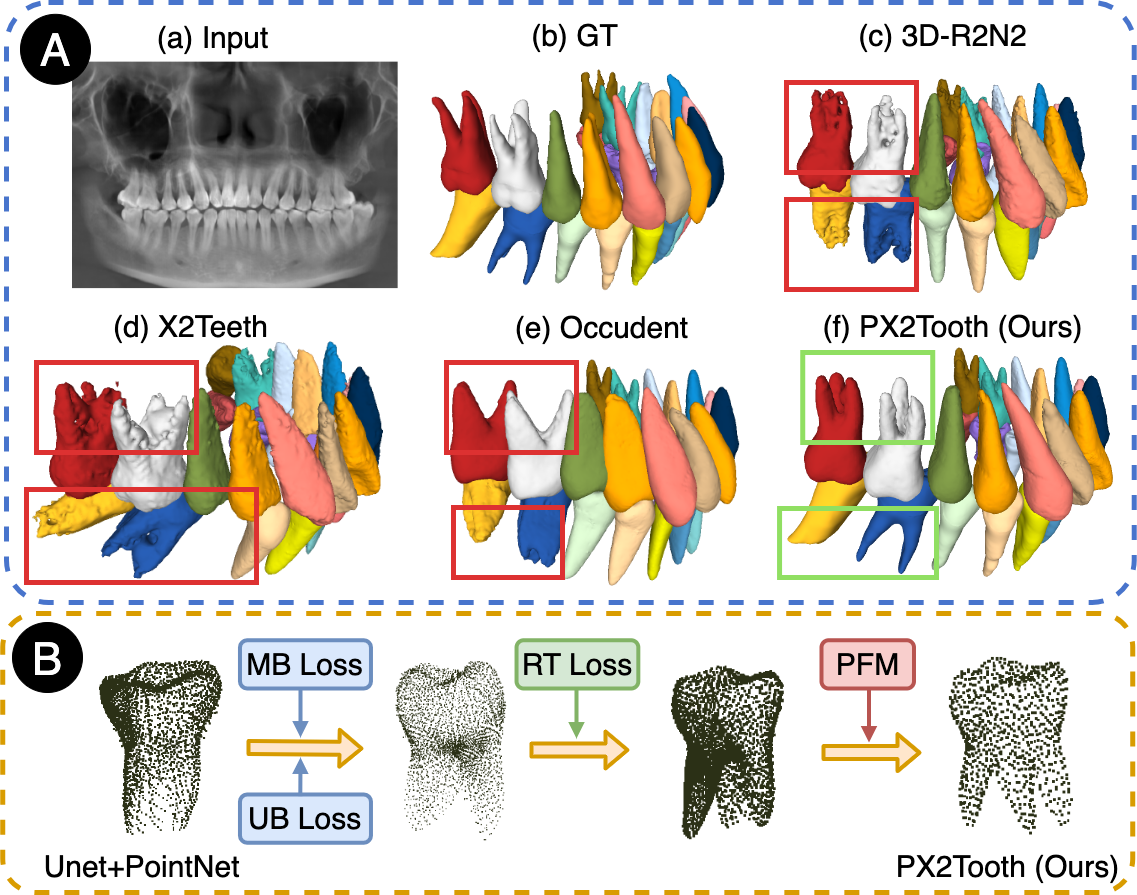}
    \caption{Visualization of reconstruction outputs. (a) overall comparison with baselines. (b) illustration of the ablation experiment.}
    \label{fig:visualize}
    \vspace{-5mm}
\end{figure}

\section{Experiments}
\subsubsection{Dataset.} 
We constructed a dataset containing 499 cases, enabling a robust and reliable evaluation that is approximately 20-fold larger than existing datasets in the field.

Firstly, we collected 499 CBCT scans from various orthodontic patients. The 3D images were resampled to the actual size of real teeth based on their respective spacing scales~\cite{ref_article24}. Then, to obtain precise 3D tooth label information, a meticulous effort was made by professional dentists to classify and finely annotate the 3D structure of individual teeth according to tooth numbers for all 499 CBCT scans. Subsequently, we employed the approach outlined by Yun et al.~\cite{ref_article15} to construct our Panoramic X-rays dataset, projecting 3D images into 2D Panoramic X-rays at a 1:1 ratio. Using the same method, we also obtained labels for the Panoramic X-rays. Finally, we partitioned the dataset into training, validation, and test sets in an 8:1:1 ratio.

\subsubsection{Implementation Details.} 

In the implementation of PXSegNet, each PX image undergoes processing with two 3x3 convolutions (unpadded), followed by ReLU activations and 2x2 max pooling for downsampling, doubling feature channels at each step. The expansive path features feature map upsampling, 2x2 up-convolutions halving feature channels, and integration with cropped maps from the contracting path plus two subsequent 3x3 convolutions with ReLU. TGNet utilizes MLPs with 1D Convolutions of 64, 128, and 1024 channels, each followed by a ReLU. PX2Tooth, trained on dual NVIDIA GeForce RTX 3090 GPUs using Adam optimizer, handles end-to-end network training with a minibatch size of 2, targeting the generation of approximately 30 teeth per PX image. Training starts with a learning rate of 1e-5, adjusted down by 0.7 every 10 epochs. Our method is cost-effective, training on a single RTX 3090 for 56.23 hours with 0.692 GB VRAM and 103.94 million parameters. It infers 32 teeth in 6.03 seconds, making it highly efficient for medical center deployment.

\subsubsection{Baseline Models.} We considered several state-of-the-art models as baselines, including 3D-R2N2~\cite{ref_article3}, PSGN~\cite{ref_article5},  Pix2mesh~\cite{ref_article11}, Occupancy Networks (OccNet)~\cite{ref_article8}, Occudent Model~\cite{ref_article2}, and X2Teeth~\cite{ref_article6}. 

The X2Teeth proposed the earliest solutions to generate 3D teeth from a single panoramic X-ray, offering a simple and straightforward groundwork for subsequent work. 
Building on the X2Teeth, the Occudent method employs neural implicit functions to evaluate if a given point in 3D space is within the tooth structure, thus implicitly delineating the contours of the tooth's 3D shape.
 
Additionally, the 3D-R2N2, PSGN, and Pix2mesh belong to general 3D generation techniques designed for widely-used 3D datasets ~\cite{ref_article14}.

\subsubsection{Main Results.} Table~\ref{tab1} demonstrates PX2Tooth's superiority over baselines in all metrics. 
Voxel-based models like 3D-R2N2 and X2Teeth capture the oral cavity's overall shape but miss detailed tooth complexities, highlighting the challenge of converting PX images into 3D structures. 
Unlike traditional encoder-decoder methods (X2Teeth, Pix2mesh, PSGN), TGNet excels in processing unordered point clouds. 
Conversely, OccNet and Occudent, which convert point clouds into discrete occupancy grids, face potential information loss and lower accuracy, showcasing TGNet's advantage in preserving detail and accuracy.

\begin{table}
\vspace{-2mm} 
\caption{Comparison with baselines. ($\downarrow$): the lower the better, ( $\uparrow$ ) : the higher the better.
MMD-CD and MMD-EMD scores are multiplied by 100 and 10, respectively.}\label{tab1}
\begin{center}
\setlength{\tabcolsep}{3.5mm}{
\begin{tabular}{l|ccc}

\toprule[1.5pt]
Method &   IoU  ( $\uparrow$ ) & MMD - CD ($\downarrow$)  & ( MMD - EMD $\downarrow$) \\
\hline
3D-R2N2 &  0.549±0.004 & 2.037±0.006   & 2.357±0.005\\
PSGN & 0.612±0.005  & 1.383±0.009   & 2.441±0.004\\
Pix2mesh & 0.642±0.005 & 1.250±0.008  & 1.535±0.005\\
X2Teeth & 0.636±0.015 & 0.792±0.003  & 1.649±0.006\\
OccNet & 0.651±0.004 & 0.802±0.005  & 1.317±0.008\\
Occudent & 0.681±0.004 & 0.671±0.004  & 1.232±0.007\\
PX2Tooth(Ours) & {\bfseries 0.793±0.004} &  {\bfseries 0.424±0.005}  &  {\bfseries 0.997±0.006}\\
\bottomrule[1.5pt]
\end{tabular}}
\end{center}
\vspace{-10mm}
\end{table}

\begin{table}
\vspace{-2mm} 
\caption{Ablation experiment comparison. ($\downarrow$): the lower the better, ( $\uparrow$ ) : the higher the better.
MMD-CD scores are multiplied by 100.}\label{tab2}
\begin{center}
\setlength{\tabcolsep}{4mm}{
\begin{tabular}{l|cc}
\toprule[1.5pt]
Method &   IoU  ( $\uparrow$ ) & MMD - CD ($\downarrow$) \\
\hline
Unet~\cite{ref_article12} + PointNet~\cite{ref_article13} &  0.63±0.004 & 1.383±0.009  \\
Unet + MB Loss + PointNet & 0.661±0.005  & 1.250±0.008   \\
Unet + UB Loss + PointNet & 0.659±0.003  & 1.290±0.008   \\
PXSegNet + PointNet & 0.692±0.005 & 1.112±0.008 \\
PXSegNet + PointNet + RT Loss  & 0.736±0.015 & 0.792±0.003  \\
PXSegNet + PointNet + PFM  & 0.744±0.004 & 0.802±0.005  \\
PX2Tooth(Ours) & {\bfseries 0.793±0.004} &  {\bfseries 0.424±0.005}   \\
\bottomrule[1.5pt]
\end{tabular}}
\end{center}
\vspace{-10mm} 
\end{table}

\begin{table}
\caption{Teeth level analysis. Average IoU statistics of each category of teeth according to FDI tooth number. Baseline refers to Unet + PointNet.}\label{tab3}
\begin{center}
\footnotesize 
\setlength{\tabcolsep}{0.3mm}{
\begin{tabular}{l|ccccccc|ccccccc}

\toprule[1.5pt]
Method &  11 & 12 & 13 & 14 & 15 & 16 & 17&  21 & 22 & 23 & 24 & 25 & 26 & 27 \\
\hline
Baseline & 0.65 & 0.63 & 0.67 & 0.62 & 0.67 & 0.60 & 0.66 & 0.59 & 0.61 & 0.65 & 0.61 & 0.69 & 0.59 &  0.65\\

PX2Tooth  & {0.78} &  { 0.78} & { 0.84} & { 0.77}& { 0.83}& {0.74}& {\bfseries  0.91} & { 0.78} &  { 0.77} & { 0.81} & { 0.76}& { 0.84}& { 0.72}& {\bfseries  0.84}   \\
\bottomrule[1.5pt]

Method &  31 & 32 & 33 & 34 & 35 & 36 & 37&  41 & 42 & 43 & 44 & 45 & 46 & 47 \\
\hline
Baseline & 0.62 & 0.65 & 0.63 & 0.61 & 0.63 & 0.58 & 0.67 &0.61 & 0.63 & 0.59 & 0.62 & 0.64 &0.58 &  0.69 \\

PX2Tooth  & { 0.83} &  { 0.86} & { 0.81} & { 0.77}& { 0.80}& { 0.75}& { \bfseries 0.86} & { 0.77} &  { 0.83} & { 0.78} & { 0.80}& { 0.79}& { 0.75}& {\bfseries 0.85} \\
\bottomrule[1.5pt]

\end{tabular}}
\end{center}
\vspace{-8mm} 
\end{table}

\noindent \textbf{Ablations.} Visualization in Fig.~\ref{fig:visualize} (b) showcases the impact of sequentially incorporating MB loss, UB loss, RT loss, and PFM on our model's performance. Table~\ref{tab2} details our ablation study results, highlighting the contribution of each component. Introducing MB Loss alone led to a 3.1\% increase in IoU while incorporating UB Loss independently resulted in a 2.9\% improvement. Adding RT Loss atop MB and UB Losses further boosted performance by 4.4\%. Including the PFM module on this foundation, our final performance surged by an additional 5.7\%, reaching a peak IoU of 79.3\%. These results underline the effectiveness of the components proposed in our method.

\subsubsection{Tooth Level Analysis.} 
\begin{figure}[htbp]
    \vspace{-5mm}
    \centering
    \includegraphics[width=0.95\linewidth]{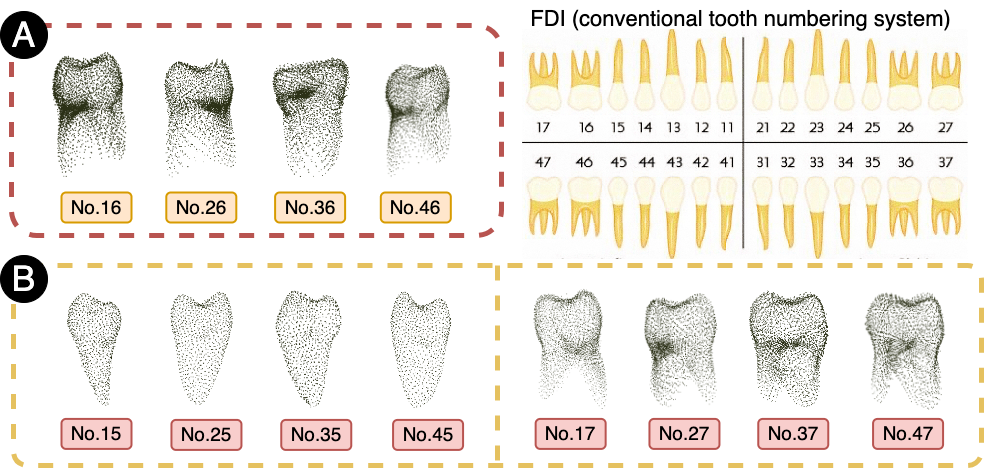}
    \caption{FDI shows the position of each tooth corresponding to the oral cavity, (a) shows the visualization problem of teeth that are not satisfactory enough (lack of root tip information). (b) shows the category tooth visualization with excellent average high IoU performance.}
    \label{fig:Number_show}
    \vspace{-5mm}
\end{figure}

Table~\ref{tab3} highlights the performance of individual tooth classes in our method, revealing that teeth numbers 17, 27, 37, and 47 excel in reconstruction due to advantageous positioning and feature richness, as shown in Fig.~\ref{fig:Number_show} (a). Teeth 15, 25, and 35 also show significant improvement, benefitting from their unique shapes. However, teeth 16, 26, 36, and 46 face minor challenges in achieving ideal outcomes, as depicted in Fig.~\ref{fig:Number_show} (b).

\noindent \textbf{Visualization.} Fig.~\ref{fig:visualize} (a) demonstrates the effectiveness of our method in producing 3D teeth meshes closely resembling the ground truth, outperforming voxel-based methods like X2Teeth and 3D-R2N2, which lose root and spatial detail and suffer from missing teeth. While Occudent improves on positional accuracy, it misrepresents tooth roots and sizes. In contrast, our approach accurately reconstructs detailed shapes and positions for various tooth types.

\subsubsection{Limitation on Clinical Validation.} Current methods primarily focus on teeth, whereas CBCT images encompass broader anatomical details such as the jawbone and nerve canals, areas which our method has not yet been validated to accurately capture. Furthermore, the clinical applicability of translating findings from 2D X-rays to 3D CBCT scans, especially in complex cases, remains to be validated. These challenges highlight the need for further research to ensure the robustness and generalizability of our approach across diverse clinical scenarios.

\section{Conclusion and Future Work}

In the paper, we proposed PX2Tooth, an end-to-end framework that accurately transforms 2D PX images into 3D tooth point clouds, bypassing the need for extra labeling. Leveraging a two-step strategy, PXSegNet segments teeth with detailed positional data, while TGNet, enhanced by a Prior Fusion Module (PFM), efficiently generates 3D teeth. We established a large-scale dataset with 499 CBCT and PX pairs to ensure reliable evaluation. PX2Tooth significantly outperforms existing methods with an IoU of 0.793, demonstrating AI's transformative potential in digital dentistry. 
Future work will aim to enhance the accuracy and efficiency of reconstruction while broadening the model's application to encompass a wider range of real-world dental situations.

\begin{credits}
\subsubsection{\ackname} This work is supported by the National Natural Science Foundation of China (Grant No. 62106222), the Natural Science Foundation of Zhejiang Province, China(Grant No. LZ23F020008), and the Zhejiang University-Angelalign Inc. R\&D Center for Intelligent Healthcare.

\subsubsection{\discintname}
Huikai Wu is employed by ChohoTech and Yang Feng is employed by Angelalign Technology Inc.
\end{credits}

\end{document}


\begin{table}
\caption{Teeth level analysis. Average IoU statistics of each category of teeth according to FDI tooth number. 11-17 represents the upper left teeth, 21-27 represents the upper right teeth, 31-37 represents the lower right teeth, 41-47 represents the lower left teeth, and the order of 1-7 is from the middle to the edge. }\label{tab1}
\begin{center}
\setlength{\tabcolsep}{1.2mm}{
\begin{tabular}{l|ccccccc}

\toprule[1.5pt]
Method &  11 & 12 & 13 & 14 & 15 & 16 & 17 \\
\hline
Unet[29]+PointNet[31] & 0.650 & 0.639 & 0.671 & 0.628 & 0.673 & 0.606 & 0.668 \\
Unet+L1+PointNet & 0.679 & 0.654 & 0.706 & 0.666 & 0.703 & 0.662 & 0.704 \\
Unet+L2+PointNet & 0.688 & 0.662 & 0.731 & 0.701 & 0.749 & 0.658 & 0.776 \\
TeethNet+PointNet & 0.701 & 0.705 & 0.768 & 0.727 & 0.776 & 0.719 & 0.828 \\
TeethNet+PointNet+L3  & 0.742 & 0.737 & 0.812 & 0.749 & 0.788 & 0.736 & 0.862 \\
TeethNet+PointNet+PFM & 0.721 & 0.746 & 0.808 & 0.745 & 0.800 & 0.739 & 0.844 \\
PX2Tooth(Ours)  & {\bfseries 0.783} &  {\bfseries 0.789} & {\bfseries 0.841} & {\bfseries 0.774}& {\bfseries 0.837}& {\bfseries 0.744}& {\bfseries 0.918}  \\
\bottomrule[1.5pt]

Method &  21 & 22 & 23 & 24 & 25 & 26 & 27 \\
\hline
Unet[29]+PointNet[31] & 0.593 & 0.616 & 0.654 & 0.610 & 0.698 & 0.592 &  0.658 \\
Unet+L1+PointNet & 0.609 & 0.629 & 0.687 & 0.633 & 0.715 & 0.623  & 0.691 \\
Unet+L2+PointNet & 0.631 & 0.638 & 0.692 & 0.649  & 0.732 & 0.651 &  0.699 \\
TeethNet+PointNet & 0.641 & 0.679 & 0.723 & 0.699  & 0.758 & 0.681 & 0.723  \\
TeethNet+PointNet+L3  & 0.688 & 0.656 & 0.764 & 0.737 & 0.788 & 0.706 & 0.764 \\
TeethNet+PointNet+PFM & 0.685 & 0.641 & 0.753 & 0.799 & 0.811 & 0.664 & 0.809 \\
PX2Tooth(Ours)  & {\bfseries 0.784} &  {\bfseries 0.770} & {\bfseries 0.810} & {\bfseries 0.762}& {\bfseries 0.841}& {\bfseries 0.720}& {\bfseries 0.842}  \\
\bottomrule[1.5pt]

Method &  31 & 32 & 33 & 34 & 35 & 36 & 37 \\
\hline
Unet[29]+PointNet[31] & 0.623 & 0.654 & 0.631 & 0.609 & 0.636 & 0.588 & 0.671  \\
Unet+L1+PointNet  & 0.658 & 0.689 & 0.690 & 0.657 & 0.662 &  0.644 & 0.701    \\
Unet+L2+PointNet  & 0.667 & 0.691 & 0.667 & 0.632 & 0.676 & 0.654 &  0.755   \\
TeethNet+PointNet &0.694 & 0.723 & 0.731 & 0.689 & 0.718 & 0.717 &  0.798 \\
TeethNet+PointNet+L3& 0.756 & 0.789 & 0.765 & 0.752 & 0.769 & 0.719 & 0.829 \\
TeethNet+PointNet+PFM & 0.774 & 0.799 & 0.778 & 0.731 & 0.788 & 0.722 & 0.834 \\
PX2Tooth(Ours)  & {\bfseries 0.833} &  {\bfseries 0.866} & {\bfseries 0.805} & {\bfseries 0.774}& {\bfseries 0.806}& {\bfseries 0.747}& {\bfseries 0.861}  \\
\bottomrule[1.5pt]

Method &  41 & 42 & 43 & 44 & 45 & 46 & 47 \\
\hline
Unet[29]+PointNet[31] &0.616 & 0.633 & 0.598 & 0.626 & 0.639 &0.589 &  0.698 \\
Unet+L1+PointNet & 0.644 & 0.665 & 0.655 &0.659 & 0.658 &0.638 &  0.732   \\
Unet+L2+PointNet & 0.661 & 0.659 &0.649 & 0.672&0.679 &0.627 &  0.769   \\
TeethNet+PointNet &0.701 & 0.697 &0.692 & 0.707& 0.729 &0.683 & 0.789  \\
TeethNet+PointNet+L3 & 0.731 & 0.783 & 0.748 & 0.739 & 0.762 & 0.733 & 0.822 \\
TeethNet+PointNet+PFM & 0.750 & 0.757 & 0.733 & 0.761 & 0.759 & 0.720 & 0.819 \\
PX2Tooth(Ours)  & {\bfseries 0.773} &  {\bfseries 0.834} & {\bfseries 0.776} & {\bfseries 0.802}& {\bfseries 0.793}& {\bfseries 0.753}& {\bfseries 0.856}  \\
\bottomrule[1.5pt]

\end{tabular}}
\end{center}
\end{table}